\documentclass[conference]{IEEEtran}
\IEEEoverridecommandlockouts
\usepackage{cite}
\usepackage{amsmath,amssymb,amsfonts}
\usepackage{algorithmic}
\usepackage{graphicx}
\usepackage{textcomp}
\usepackage{xcolor}
\usepackage{algorithm}
\usepackage{tabularx,booktabs}
\usepackage[flushleft]{threeparttable}
\usepackage{threeparttable}
\usepackage{hyperref}

\def\BibTeX{{\rm B\kern-.05em{\sc i\kern-.025em b}\kern-.08em
    T\kern-.1667em\lower.7ex\hbox{E}\kern-.125emX}}

\usepackage{amsthm}

\begin{document}

\title{Federated Loss Exploration for
Improved Convergence on Non-IID Data\\
\thanks{\copyright~2024 IEEE. This is the author's accepted manuscript of an article published in the \textit{Proceedings of the 2024 International Joint Conference on Neural Networks (IJCNN 2024)}. The final version of record is available at \href{https://doi.org/10.1109/IJCNN60899.2024.10651455}{https://doi.org/10.1109/IJCNN60899.2024.10651455}.}
}




\author{
\IEEEauthorblockN{Christian Internò\IEEEauthorrefmark{1}, Markus Olhofer\IEEEauthorrefmark{2}, Yaochu Jin\IEEEauthorrefmark{3}, Barbara Hammer\IEEEauthorrefmark{1}}

\IEEEauthorblockA{
\IEEEauthorrefmark{1}\textit{CITEC, Bielefeld University}, Bielefeld, Germany\\
\IEEEauthorrefmark{2}\textit{Honda Research Institute Europe}, Offenbach am Main, Germany\\
\IEEEauthorrefmark{3}\textit{School of Engineering, Westlake University}, Hangzhou, China
}

\IEEEauthorblockA{
\{christian.interno@, bhammer@techfak.\}uni-bielefeld.de, Markus.Olhofer@honda-ri.de, jinyaochu@westlake.edu.cn
}
}


\maketitle
\thispagestyle{plain}

\begin{abstract}
Federated learning (FL) has emerged as a groundbreaking paradigm in machine learning (ML), offering privacy-preserving collaborative model training across diverse datasets. Despite its promise, FL faces significant hurdles in non-identically and independently distributed (non-IID) data scenarios, where most existing methods often struggle with data heterogeneity and lack robustness in performance. This paper introduces Federated Loss Exploration (FedLEx), an innovative approach specifically designed to tackle these challenges. FedLEx distinctively addresses the shortcomings of existing FL methods in non-IID settings by optimizing its learning behavior for scenarios in which assumptions about data heterogeneity are impractical or unknown. It employs a federated loss exploration technique, where clients contribute to a global guidance matrix by calculating gradient deviations for model parameters. This matrix serves as a strategic compass to guide clients' gradient updates in subsequent FL rounds, thereby fostering optimal parameter updates for the global model. FedLEx effectively navigates the complex loss surfaces inherent in non-IID data, enhancing knowledge transfer in an efficient manner, since only a small number of epochs and small amount of data are required to build a strong global guidance matrix that can achieve model convergence without the need for additional data sharing or data distribution statics in a large client scenario. Our extensive experiments with state-of-the art FL algorithms demonstrate significant improvements in performance, particularly under realistic non-IID conditions, thus highlighting FedLEx's potential to overcome critical barriers in diverse FL applications.\\
Implementation code:\href{https://github.com/ChristianInterno/FedLEx}{https://github.com/ChristianInterno/FedLEx}.
\end{abstract}


\section{Introduction}

\IEEEPARstart{In}{The present} age of technology, both individuals and institutions play a key role in generating and managing data. The exponential growth in the generation of data poses significant obstacles for ML, especially for intricate data-driven deep learning (DL) models. These challenges primarily revolve around the distribution of data and the increasing concerns regarding privacy issues associated with personal data. The streaming, collection, and storage of large data sets can become impractical, making it challenging to process them further. Additionally, traditional DL methods rely on centralized data collection for training the models.

The privacy issue regarding data collection and the regulations imposed by governments, such as the European General Data Protection Regulation (GDPR), have brought attention to FL as a viable approach to training DL models that can address these limitations \cite{mcmahan2023communicationefficient}.

FL facilitates parallel training among various participating clients, effectively mitigating privacy risks by negating the need for data sharing. The main orchestration of this process is carried out by a global model hosted on a server. This model collects and combines only locally trained client models, ensuring a secure learning environment.

In particular, the \emph{ De Facto} standard of FL \texttt{FedAvg} \cite{mcmahan2023communicationefficient}, operates on a unique collaborative principle. In this approach, the global model distributes a model to various clients for localized training. Each client's contribution to the global model update is proportional to the size of its dataset, ensuring that clients with larger datasets exert a more significant influence on the ultimate global model. This is achieved by computing the parameters as a weighted average of the individual parameters learned by each client. The process, characterized by this collaborative learning dynamic, is iteratively repeated until the model converges to an optimal state. Empirical evidence has demonstrated the stability of this approach, even when dealing with non-convex optimization problems. Consequently, it is commonly employed as a benchmark for comparing newly developed FL protocols.

Despite the considerable promise of FL, its effectiveness is significantly hindered by the prevalence of non-IID data in real world scenarios \cite{zhu2021federated, karimireddy2021scaffold}. In FL, 'non-IID data' refer to the unique statistical properties of each client's dataset, reflecting the varied environments of their origin. 

Such variations pose a substantial challenge to algorithms such as \texttt{ FedAvg}, which are designed for more homogeneous data distributions. This situation creates conflicting training goals for the local and global models, and frequently leads to the convergence towards different local optima. This leads to biased client model updates for the global model, hindering the global convergence and skews the performance. Discrepancies exacerbate generalization errors and training divergence, especially when aggregated model updates fail to align with the optimal convergence direction \cite{zhu2021federated, karimireddy2021scaffold}.

This divergence underscores the need for innovative personalized strategies in FL, particularly in data handling and model training, to improve the generalizability and reliability of FL models across diverse data landscapes.

These strategies aim to personalize the learning process to the diverse data distributions and requirements between clients, ensuring more efficient results. In general, there are two distinct approaches to personalization in FL:
\textbf{1) Global model personalization:} Here, the focus is on maintaining a single global model, applicable to all clients. The challenge lies in how to integrate diverse client updates into the global model while mitigating biases caused by non-IID data. Techniques such as selective sharing of homomorphic encrypted \cite{yan2023hedksap} client data, use of additional statistical information on clients, and data augmentation \cite{10214273} are used to create a more representative unbiased global model. Works such as \cite{Tan_2022, mcmahan2023communicationefficient, karimireddy2021scaffold, 9084352} provide insight into these approaches.
\textbf{2) Client models personalization:} This strategy diverges significantly from the first by advocating the development of customized models for each client. Instead of a one-size-fits-all model, this approach allows models that are specifically tuned to the data characteristics of each client, enhancing relevance and performance. It includes designing personalized client models, by modifying the topology of neural networks for individual clients, leveraging similarities among client data to guide model personalization and the use of multiple global models as explored in studies such as \cite{Tan_2022, 9494459, zhu2021federated, fallah2020personalized}.

Although global model personalization in FL aims to develop an optimal single robust model applicable across all clients, it may fail to address the unique requirements of individual clients. In contrast, personalizing client models offers specificity to individual needs, providing customized solutions that a universal model may miss. However, this approach introduces greater computational and infrastructural complexity and moves away from the ideal of a single, global model-based FL setting. Therefore, choosing between these approaches requires a careful evaluation of the trade-offs: the need for improved accuracy, the complexities involved, and the priorities of the specific real-world scenario.

Given these considerations, the development of an innovative FL schema is crucial. These schema should effectively achieve three key objectives:
\textbf{(i)} Improve global aggregation strategies in FL while maintaining strict privacy standards. 
\textbf{(ii)} The adaptability of these schema is important in achieving a balance between meeting the unique learning needs of each client and maintaining the efficiency and universality of a single global model. 
\textbf{(iii)} Furthermore, the schema should be formulated in such a way that it can easily integrate with established FL aggregation algorithms, thus improving performance even further.

In our research, we delve into the integration of knowledge transferring of aggregated gradient behavior data of various clients to address the challenges posed by the intricate and highly dimensional loss functions characteristic of deep neural networks. Traditional gradient descent methods, commonly used in DL, struggle in these settings. Their limitation lies in a lack of adaptability to complex loss surfaces that vary significantly between different clients, leading to biased weight updates and inefficient convergence. Additionally, Zhang et al. \cite{9671693} demonstrate that a critical issue that affects the accuracy is the large gradient diversity of the different client models. 

Our proposed method seeks to remedy this by enabling clients to efficiently navigate these diverse loss landscapes. This is achieved by sharing knowledge through adaptive gradient guidance parameters, which not only facilitate more effective client updates but also assist the global model in converging faster. Such an approach holds promise for enhancing the robustness and effectiveness of FL systems, particularly in complex DL tasks across various sectors.

We assume a scenario focused on improving learning sessions for individual client models within strongly non-IID environments by aggregating additional secure gradient information from other clients, rather than relying solely on weight updates. 
In our assumed single-server architecture, each client operates on the same deep neural network structure, but with varying class distributions in their tasks.

In the field of FL, it has been common practice to use knowledge transfer methods \cite{trflearn} as a strategy to enhance model performance. Typically, these methods have primarily focused on weight-sharing of large pre-trained models to improve clients performance through fine-tuning \cite{zhu2021federated}. However, it often comes with its own set of limitations, and they often require a large dataset and high computational resources \cite{trflearn}.

Our work acknowledges this traditional knowledge transfer in FL, but diverges significantly in approach. We introduce a novel method that has not yet been explored in the context of FL. We propose the Federated Loss Exploration (\texttt{FedLEx}), introducing the concept of guided transfer learning \cite{nikolić2023guided} within an FL framework, which uses gradient behavior analysis to guide the learning process.
Unlike traditional FL aggregation methods, \texttt{ FedLEx} does not rely solely on weight sharing. Instead, it embarks on a new path by focusing on the aggregation gradient information from different loss surfaces of clients. This approach is particularly tailored to address the challenges in strongly non-IID scenarios, where the bias introduced due to varied class distributions across clients significantly hinder learning efficacy.

\textbf{The primary contributions} of this work can be summarized as follows: \textbf{1)} We propose a new FL algorithm called \texttt{FedLEx} specifically designed for non-IID scenarios. \textbf{2)} We illustrate the adaptability and compatibility of \texttt{FedLEx} with popular FL aggregation algorithms, highlighting its usefulness in a wide range of FL settings. \textbf{3)} Our empirical experiments encompass various benchmark datasets, tasks, and realistic non-IID scenarios, providing strong evidence of the effectiveness of \texttt{FedLEx}. \textbf{4)} Through detailed ablation studies, we analyze the individual and collective impact of the components of \texttt{FedLEx}, demonstrating its efficiency and adaptability.

\section{RELATED WORKS} 

To address non-IID challenges, transfer learning and knowledge distillation have gained popularity in FL. The core idea is to transfer knowledge from the server or other clients to improve clients performance on unknown data. Techniques such as adapting large pre-trained models to local models in FL are prevalent \cite{li2019fedmd, wang2019federated, lin2021ensemble}. Furthermore, adaptations of model-agnostic meta-learning, such as the \texttt{Reptile} algorithm \cite{DBLPdfgh}, and multitask learning approaches such as \texttt{ MOCHA} \cite{fallah2020personalized}, have shown promise. Nevertheless, such algorithms may face challenges with convergence when dealing with large-client scenarios as a result of sequential fine-tuning \cite{zhu2021federated}.

Some researchers have turned to hierarchical clustering \cite{sattler2019clustered} to categorize clients based on weight update patterns, thus facilitating effective grouping of clients. Others have explored identifying global data distributions that more accurately reflect the diversity of client data, moving beyond traditional weighted mean approaches to improve model suitability \cite{9174245}. However, these methods can introduce new challenges related to data security because of their reliance on data similarity.

Data sharing is another method to handle non-IID data and local data imbalance\cite{zhu2021federated}. Data augmentation methods such as vanilla, mix-up, and GAN-based techniques have been utilized in FL \cite{zhu2021federated,10214273}. While straightforward and effective, they often conflict with FL's privacy-preserving principles by potentially requiring to share local data with the server. 

Closely related to our work are advances in adaptive gradient methods for training. Adaptive optimizers, such as Adagrad and Adam, are gradually replacing the conventional SGD method in FL. Techniques such as group normalization \cite{wu2018group}, which compute per-group statistics by partitioning channels of training data, have proven effective in accelerating convergence \cite{9671693}. \texttt{FedProx} \cite{li2020federated}, an innovative method, focuses on varying local updates before convergence, allowing participants to adapt the global model based on their update disparities. This personalization adds a valuable dimension to the training process. However, concerns remain about the sensitivity of local moving statistics to client data distribution and the potential failure of aggregated global statistics to converge on non-IID data. Additionally, such algorithms often lack foresight into the loss function landscape, which can lead to being trapped in steep local minima. To the best of our knowledge, no existing work explore transferring knowledge about the loss function map to guide and adapt client gradients in an FL setting, a gap our research aims to address.

\section{PROPOSED METHOD}
\subsection{Notations}
We first define the parameter settings required for \texttt{FedLEx}: a total number of communication rounds (\( R \)), a local batch size (\( B \)), a number of local epochs (\( E \)), a total number of clients (\( C \)), and a fraction of clients selected at each round (\( K \)). Additionally, two hyperparameters are required for our proposed method: the fraction of clients selected to perform the exploration phase (\( C_{exp} \)), and the number of exploration epochs (\( E_{exp} \)).

\subsection{Problem Statement and Objectives}

In FL, our challenge is to train a global model across \( C \) clients (\( c_i \)), each with a unique dataset \( D_i = \{ (x_i, y_i) \} \) from diverse non-IID distributions \( P_i \), impacting learning patterns and the global model performance. The aim is to develop a consistent global model that preserves privacy and prevents data sharing between clients, with each client's model \( c_i \) defined by parameters \( W_i \) and a deep neural network.

We focus on optimizing the model to handle non-IID data disparities and improve convergence using a new aggregation strategy, integrating guided transfer learning with a federated loss exploration. The expected loss for each client's model is:
\begin{equation}
    L_{P_i}(h_i, W_i) = \mathbb{E}_{(x_i, y_i) \sim P_i} [l(h_i(x_i; W_i), y_i)],
\end{equation}
where \( L_{P_i} \) represents the expected loss over data distribution \( P_i \) for client \( c_i \) with hypothesis \( h \) from space \( H \) which encompasses all the potential models. These models could be generated within the constraints of the chosen deep neural network architecture and model parameters \( W_i \).

The goal is to align global model parameters \( W_{global} \) with the optimal \( W_{global}^* \) that minimizes the aggregated loss:
\begin{equation}
    \min_{h_i \in H} \frac{1}{C} \sum_{i=1}^{C} L_{P_i}(h_i, W_{i,t})
\end{equation}
with \( W_{i,t} \) being the parameters of client \( c_i \) at iteration \( t \).

\texttt{FedLEx} introduces \( G_{\text{global}} \), gradient guidance parameters from a federated loss exploration. Clients adapt their weight updates as \( \Delta W_{i, \text{modulated}} = \Delta W_i \times G_{\text{global}} \), aligning learning with collective goals and aiding efficient convergence.

The  \( G_{\text{global}} \) minimizes variance in client weight updates in an meaningfully non random way:
    \(\sigma^2_{\Delta W} = \text{Var}[\Delta W_1, \Delta W_2, \ldots, \Delta W_C] \),
where \( \sigma^2_{\Delta W} \) is the variance in updates \( \Delta W_i \). With \( G_{\text{global}} \), we aim to reduces this variance in a meaningfully way, crucial for a efficient global convergence.

\subsection{Overview}

\begin{figure*}[t]
    \centering
    \includegraphics[width=\linewidth]{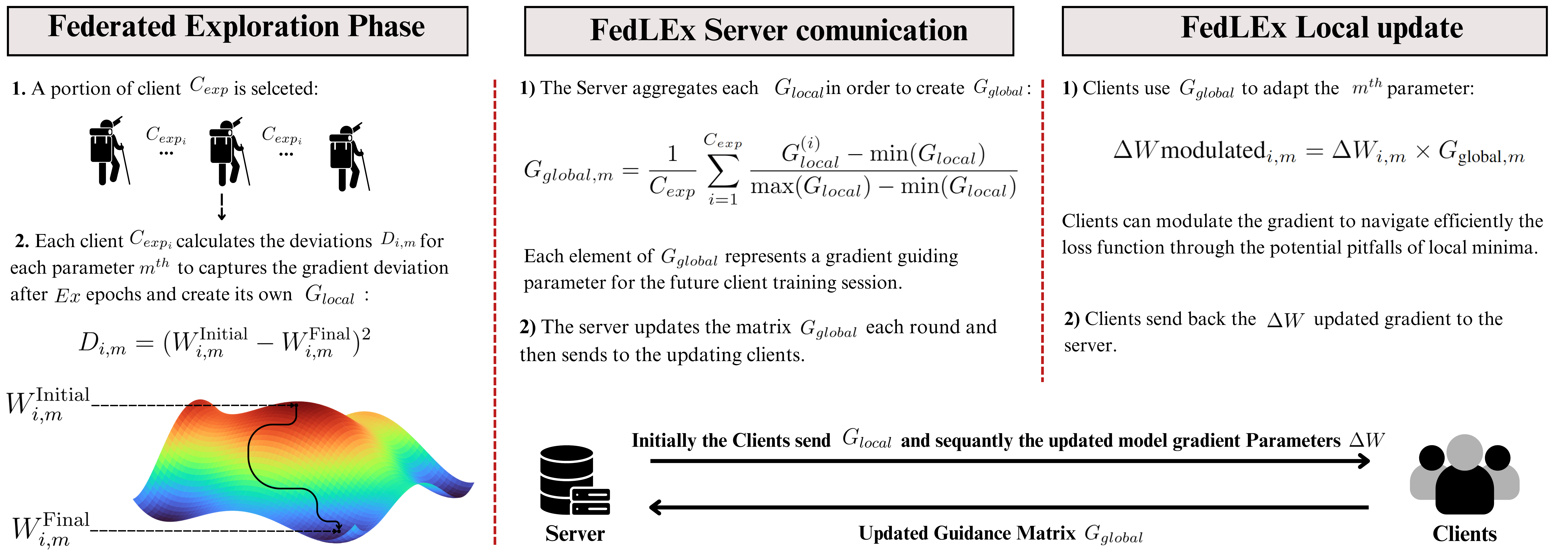}
\caption{Left: Exploration phase yields parameter deviations and local guidance matrices. Center: These matrices are aggregated into \(G_{global}\) during server communication. Right: \(G_{global}\) informs local updates, sent back to the server.}

    \label{fig:Fig-schema}
\end{figure*}

In \texttt{FedLEx}, alongside the model initialization phase, clients embark on an additional novel loss federated exploration phase to explore the surface of their own loss function. This exploration enables the incorporation of important additional gradient knowledge into the Global Guidance Matrix ($G_{global}$). This matrix is then shared among clients, serving as a compass during the future training session to guide clients and facilitate more optimized parameter updates. The overall architecture schema of \texttt{FedLEx} is depicted in Figure \ref{fig:Fig-schema}.
 
To create $G_{global}$, the selected \( C_{exp} \) portion of clients has to calculate the deviation for each gradient model parameter during the loss exploration. Specifically, the deviation $D$ for the \( m^{th} \) parameter of client model  \( C_{exp_i} \) is defined as the square difference between the initial $W_{i,m}^{\text{Initial}}$ and final $W_{i,m}^{\text{Final}}$ parameter values after \( E_{exp} \) exploration epochs:
\begin{equation}
    D_{i,m} = (W_{i,m}^{\text{Initial}} - W_{i,m}^{\text{Final}})^2.
    \label{eq:dev}
\end{equation}

This deviation from equation \ref{eq:dev} reflects the variability of the gradient encountered during the exploration of the loss landscape. It can be interpreted as an indicator of the unpredictability of the gradient or "surprise". By utilizing these deviations, \( C_{exp} \) clients share their own local matrix $G_{local}$ where each element indicates the gradient deviation value for a given model parameter.
The server aggregates these $G_{local}$ to generate \( G_{global} \) using the following normalization:
\begin{equation} 
G_{global,m} = \frac{1}{ C_{exp} } \sum_{i=1}^{ C_{exp} } \frac{G_{local}^{(i)} - \min(G_{local})}{\max(G_{local}) - \min(G_{local})} 
\label{eq:norm}
\end{equation}

Each element of \( G_{global} \) represents an average normalized deviation of the gradient for each model parameter \( m^{th} \) across the loss exploration, scaled between 0 and 1 with equation \ref{eq:norm}. 
A value near 0 implies a lower deviation, indicating a relatively stable gradient, associated with a flatter region in the loss landscape. In contrast, a value approaching 1 denotes a higher gradient deviation, suggesting a more volatile gradient that might correspond to a more steep and interesting loss region. 

Leveraging the insights derived from \( G_{\text{global}} \), $K$ clients participating in the updating round adapt the \( m^{th} \) gradient:
\begin{equation} 
 \Delta W\text{modulated}_{i,m} = \Delta W_{i,m} \times G_{\text{global},m},
 \label{eq:norm}
\end{equation}
where \( \Delta W\text{modulated}_{i,m} \) denotes the modulated update for the \( m^{th} \) parameter of client \( i \). On the other hand, \( \Delta W_{i,m} \) signifies the original gradient update for the \( m^{th} \) model parameter of client \( i \) before the use of $G_{global}$. Following this, the server receives \( \Delta W\text{modulated}_{i,m} \) to update the global model.

\subsection{Client Update}

In FedLEx, the process of updating local models is enriched using the knowledge shared through \( G_{\text{global}} \). For each client \( K_i \) that participates in the FL round update, working with a dataset \( D_i \), the local update process is guided by the gradient of the local loss function with respect to its model parameters \( W_i \), denoted as \( \nabla L_{\text{local}}(W_i) \). However, instead of using this gradient directly for the update, \texttt{FedLEx} incorporates \( G_{\text{global}} \) to modulate the gradient.
This modulation is executed as follows in the updated model parameters for each client \( K_i \):
\begin{equation}
 W\text{modulated}_i = W_i - \eta \left( \nabla L_{\text{local}}(W_i) \times G_{\text{global}} \right).
 \label{eq:locup}
\end{equation}
In equation \ref{eq:locup}, \( \eta \) is the learning rate. This process integrates the shared global knowledge encapsulated in \( G_{\text{global}} \) into the local update mechanism. The elements in \( G_{\text{global}} \) influence the magnitude of gradient updates: The parameters corresponding to lower values in \( G_{\text{global}} \) undergo smaller updates. This approach provides an implicit, informed adaptation for the gradient, aligning local updates more closely with the collective insights of all participating clients. For a detailed process of the \texttt{FedLEx} client-side mechanism, refer to Algorithm \ref{alg:clients}.

\begin{algorithm}
\caption{Client-side Procedure of \texttt{FedLEx}}
\begin{algorithmic}[1]
\STATE Client computes \(G_{\text{local}}\) to then send it to the server 
\STATE Client receive \(G_{\text{global}}\) from the server
\FOR{each communication round \( r \)}
    \STATE Client receives updated \( G_{\text{global}} \) from server
    \STATE Client computes update:\\
     \(W\text{modulated}_i = W_i - \eta \left( \nabla L_{\text{local}}(W_i) \times G_{\text{global}} \right) \)
    \STATE Client sends updated model weights \( W_i \) to server
\ENDFOR
\end{algorithmic}
\label{alg:clients}
\end{algorithm}

\subsection{Communication with the Server}
During the initial rounds, a set of \( C_{exp} \) client computes and shares the \( G_{\text{local}} \) matrices. The server aggregates these \( G_{\text{local}} \) to compute \( G_{\text{global}} \) through equation \ref{eq:norm}.

In each subsequent round, the server updates \( G_{\text{global}} \) utilizing the already stored \( G_{\text{local}} \) from a selected fraction \(K\) participating clients that have also been in the selected fraction \( C_{exp} \), and broadcasts the updated \( G_{\text{global}} \) to all clients $K$ to guide their gradient updates.

The server, upon aggregating the received model updates, proceeds to update the global model. Once updated, the global model is either ready for the next communication round or deemed ready for deployment if the convergence criteria are satisfied. For a detailed process of the \texttt{FedLEx} server communication process, refer to Algorithm \ref{alg:server}.

\begin{algorithm}
\caption{Server-side Procedure of \texttt{FedLEx}}
\begin{algorithmic}[1]
\STATE Server initializes global model and local models
\FOR{\( r = 1 \) to \( R \)}
    \IF{\( r = 1 \)}
        \STATE Server selects \( C_{exp} \) fraction of clients for initial \( G_{\text{local}} \) computation
        \STATE Receive \( C_{exp} \) different \( G_{\text{local}} \) 
        \STATE Compute initial \( G_{\text{global}} \) with equation \ref{eq:norm}
    \ELSE
        \STATE Update \( G_{\text{global}} \) with \(G_{\text{local}} \) of \(K\) participating clients that have been also in the selected fraction \( C_{exp} \)
        \STATE Send the updated \( G_{\text{global}} \) to $K$ Clients 
    \ENDIF
    \STATE Receive model updates \( \Delta W_{i} \) from each  \( K \) clients

\ENDFOR
\end{algorithmic}
\label{alg:server}
\end{algorithm}

\subsection{Computational Complexity}
The computational and communication complexities of \texttt{FedLEx} are pivotal for evaluating its feasibility and efficiency. The computational complexity mainly involves two phases: the exploration phase and the updating phase.

1. \textit{Exploration phase}: Each client \( C_i \) during the exploration phase computes the deviation for each model parameter. This has a computational complexity of \( \mathcal{O}(m \cdot E_{exp}) \), where \( m \) denotes the number of model parameters and \( E_{exp} \) the number of exploration epochs.
2. \textit{Updating Phase}: In the updating phase, clients calculate the gradient of the loss function and modulate it using \( G_{\text{global}} \), resulting in a complexity of \( \mathcal{O}(m \cdot R) \), with \( R \) representing the number of communication rounds. 
Thus, the total computational cost for each client aggregates to \( \mathcal{O}(m \cdot (E_{exp} + R)) \).

Communication complexity involves the initial exchange of \( G_{\text{local}} \) matrices from \( C_{exp} \) clients and the subsequent round-wise communication of \( G_{\text{global}} \) and model updates. Initially, the server's communication with \( C_{exp} \) clients incurs a complexity of \( \mathcal{O}(C_{exp} \cdot m) \). In each round \( R \), the server updates and communicates \( G_{\text{global}} \) to \( K \) clients and receives their model updates, leading to \( \mathcal{O}(K \cdot m \cdot R) \).
The total complexity for \( R \) rounds is a combination of the initial exploration phase and the iterative communication between server and clients:
\( \mathcal{O}(C_{exp} \cdot m) + \mathcal{O}(K \cdot m \cdot R) \).
This summarizes \texttt{FedLEx}'s resource needs, from setup to iterative execution.

\section{EXPERIMENTS}

\subsection{Experimental Setup}
Our experiments emphasize the global model performance across various realistic non-IID scenarios, drawing inspiration from and replicating the environment and experiments outlined in \cite{Hahn2022}. These evaluations are conducted on extensive benchmark datasets, providing comprehensive insights into the robustness and adaptability of our approach.

Unless stated otherwise, our standard settings for all experiments involve using a fraction of $Cs$ modulated to have five participating clients in every communication round, a batch size $B=50$, a learning rate $ \eta=0.0003$ and a total of communication rounds $R=500$ and $C=20, 100, 200$.  We employ the SGD optimizer and a weight decay factor of 0.0001.
Regarding the two FedLEx hyperparameters, we employ a number of exploring clients $C_{exp}=20$ and a maximum of exploration epochs $E_{exp}=150$.  

We implement \texttt{FedLEx} alongside a few popular state-of-the-art
FL aggregation methods. Starting from \texttt{FedAvgM} \cite{hsu2019measuring} that relies on a mitigation strategy via server momentum, \texttt{FedSgd} \cite{mcmahan2023communicationefficient} for a straightforward approach where clients compute gradients based on their local data and immediately send these gradients to the server for aggregation, \texttt{FedOpt} \cite{reddi2021adaptive} that additionally optimizes learning rates based on the estimates of the first and second moments of the gradients with the Adam algorithm, \texttt{FedProx} \cite{li2020federated} that introduces a proximal term to the local optimization problem, allowing clients to solve a personalized version of the learning task. 

Clients store their data in training and test sets, in an 80-20 split. Performance is evaluated on each client's test set, measuring top-1 accuracy metrics. In order to have a statistically robust evaluation of the performance of the FL algorithms, we repeated each experiment 20 times with different random seeds and model's weights initialization in order to evaluate the results' statistical significance. 
The FedLEx implementation code is available on \href{https://github.com/ChristianInterno/FedLEx}{https://github.com/ChristianInterno/FedLEx}.

\subsection{Experimental Results}
\textbf{Pathological Non-IID} This setting is characterized by clients having data from only two classes within a multiclass dataset. We used the MNIST \cite{6296535} and CIFAR10 \cite{articleCIFAR} dataset. We used a fully connected two-layer network for MNIST and a two-layer CNN for CIFAR10, as suggested in \cite{mcmahan2023communicationefficient}. 

In the table \ref{tab:table1}, it is possible to observe results and standard deviation for pathological non-IID experiments for top-1 accuracy.  The \texttt{FedLEx} variants outperform the traditional version in all client numbers \( C\), with the standard deviation also being generally lower for \texttt{FedLEx}, indicating better stability, especially in scenarios with a higher number of clients $C$.

\begin{figure}[t]
    \centering
    \includegraphics[width=\linewidth]{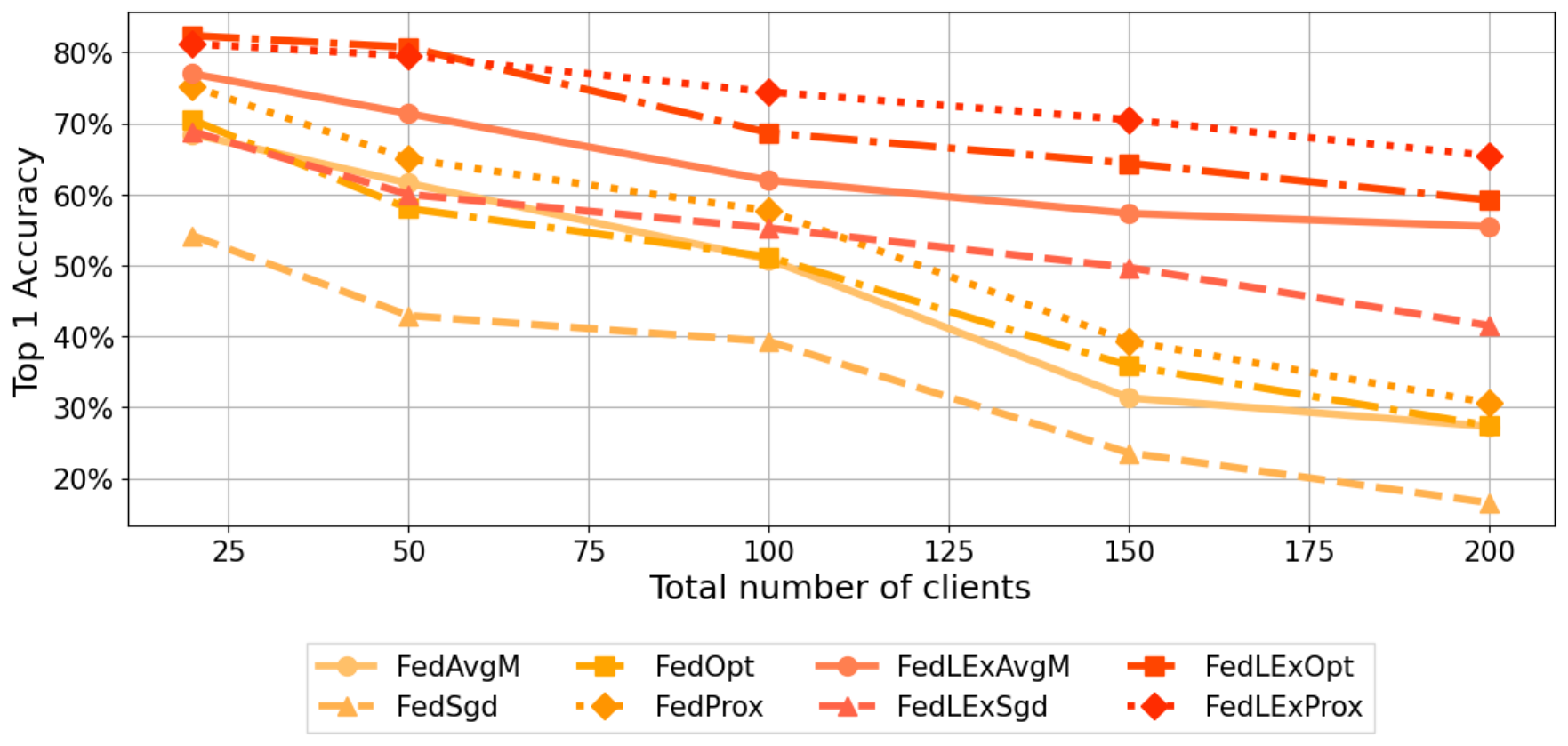}
    \caption{Performance comparison of the traditional FL algorithm and \texttt{FedLEx} variants with an increasing number of participating clients $C$.}
    \label{fig:clioentsdist}
\end{figure}

Figure \ref{fig:clioentsdist} provides a representation of the impact of the described pathological non-IID data setting on the performance of FL algorithms as the number of clients $C$ increases. It is worth noting that traditional FL algorithms represented in orange experience a noticeable decline in performance as $C$ grows, indicating their vulnerability to this challenging scenario. By contrast, the \texttt{FedLEx} variants, depicted in red, exhibit a remarkable resilience, showing only a marginal decrease in performance. This underscores the robustness of \texttt{FedLEx} variants in the face of increased client participation, making it an appealing choice to achieve a fairer FL scenario.

\begin{figure}[t]
    \centering
    \includegraphics[width=\linewidth]{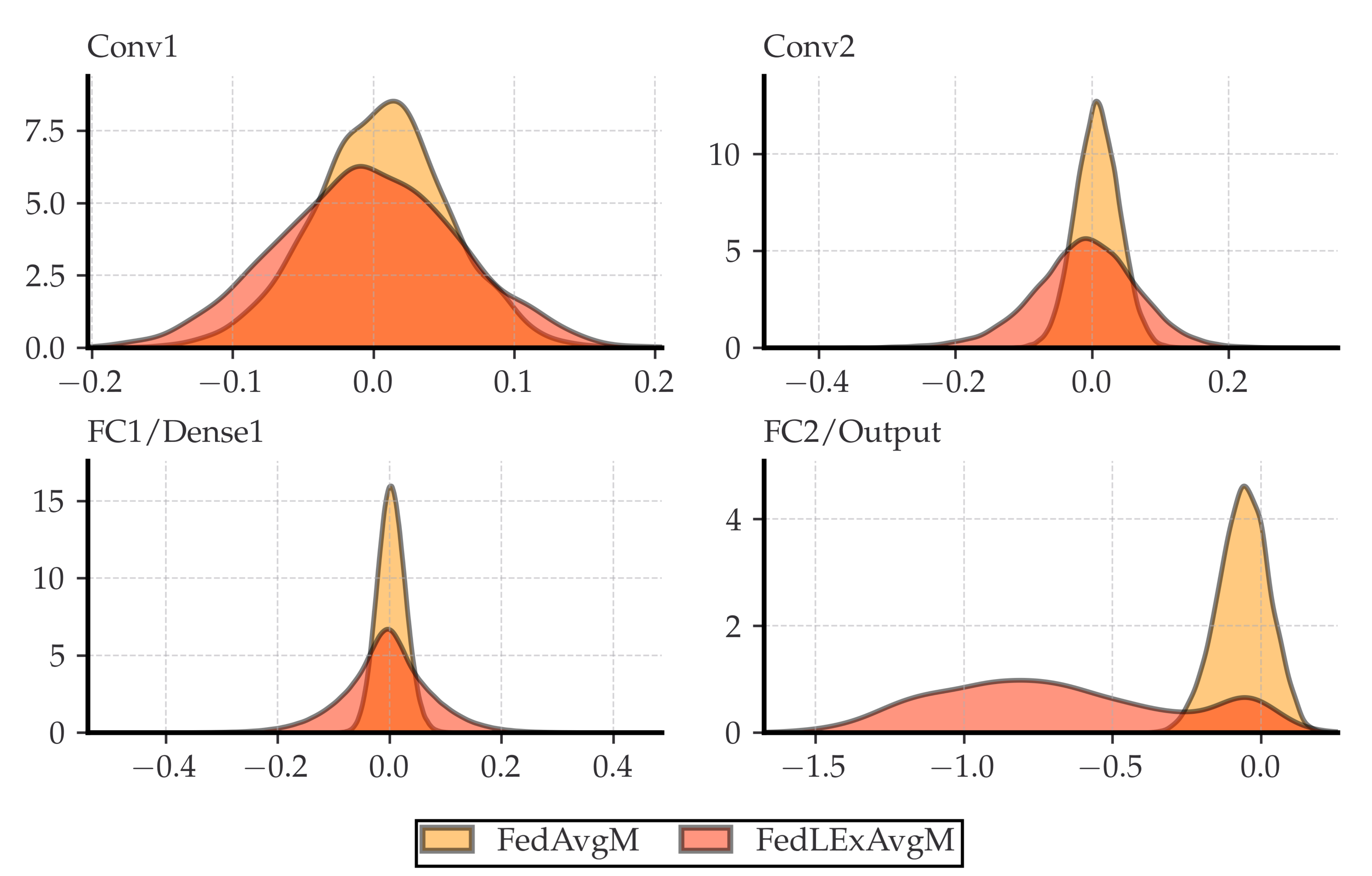}
    \caption{Comparison of kernel density estimates of the model parameter distribution with \texttt{FedLEx} implemented on \texttt{FedAvgM}.}
    \label{fig:distri}
\end{figure}

Figure \ref{fig:distri} offers a detailed visual representation of how the global guidance matrix influences the distribution of parameters within the client's deep neural network. It specifically shows the kernel density estimates of the model parameters' distribution across each layer of the TwoCNN, applied to the non-IID CIFAR10 dataset for \texttt{FedAvgM} and \texttt{FedLExAvgM}. Across all layers, there is a significant shift in the distribution of parameter values, particularly noticeable in the final fully connected layer. Each client in \texttt{FedLEx} experiences similar modifications, resulting in a substantial decrease in the variance ( \(\sigma^2_{\Delta W}\)) of weight updates for the global model. The reduction in variance helps to alleviate the bias caused by the non-IID setting, thus promoting better global convergence.


\newcolumntype{C}{>{\centering\arraybackslash}X}  
\begin{table}[t]
    \centering
    \scriptsize
    \begin{threeparttable}
        \caption{Results for pathological non-IID setting, detailing top-1 accuracy (\%) with standard deviation.}
        
        \label{tab:table1}
        \begin{tabularx}{\linewidth}{@{} l *{6}{C} @{}}
            \toprule
            Dataset & \multicolumn{3}{c}{MNIST} & \multicolumn{3}{c}{CIFAR10} \\
            \cmidrule(lr){2-4} \cmidrule(lr){5-7}
            \( C \) & 20 & 100 & 200  & 20 & 100 & 200 \\
            \midrule
            
            FedAvgM\cite{hsu2019measuring} & \( 78.50 \pm 6.75 \) & \( 60.85 \pm 6.00 \) & \( 37.30 \pm 5.80 \) & \( 52.50 \pm 4.20 \) & \( 40.60 \pm 4.75 \) & \( 35.50 \pm 5.60 \) \\

            FedSgd\cite{mcmahan2023communicationefficient} & \( 64.23 \pm 6.02 \) & \( 49.30 \pm 6.48 \) & \( 26.56 \pm 5.57 \) & \( 31.06 \pm 3.68 \) & \( 29.19 \pm 4.45 \) & \( 26.18 \pm 5.56 \) \\
            
            FedOpt \cite{reddi2021adaptive} & \( 80.51 \pm 7.80 \) & \( 61.21 \pm 2.53 \) & \( 37.41 \pm 2.13 \) & \( 59.18 \pm 7.27 \) & \( 45.81 \pm 3.05 \) & \( 28.31 \pm 2.41 \) \\

            FedProx\cite{li2020federated} & \( 85.23 \pm 7.32 \) & \( 67.70 \pm 4.16 \) & \( 40.64 \pm 3.55 \) & \( 65.64 \pm 5.97 \) & \( 44.36 \pm 2.45 \) & \( 28.33 \pm 1.64 \) \\

            \midrule

            FedLExAvgM & \( 87.00 \pm 7.10 \) & \( 72.00 \pm 6.40 \) & \( 65.50 \pm 5.70 \) & \( 72.80 \pm 4.50 \) & \( 61.80 \pm 4.90 \) & \( 57.00 \pm 5.70 \) \\

            FedLExSgd & \( 78.86 \pm 7.69 \) & \( 65.01 \pm 6.95 \) & \( 51.28 \pm 2.69 \) & \( 33.16 \pm 4.78 \) & \(32.33 \pm 7.99 \) & \( 29.88 \pm 6.50\) \\
            
            FedLExOpt & \( \textbf{92.37} \pm \textbf{4.00} \) & \( 78.74 \pm 2.65 \) & \( 69.69 \pm 3.99 \) & \( 77.42 \pm 8.37 \) & \( 64.48 \pm 3.09 \) & \( 59.58 \pm 4.17 \) \\

            FedLExProx & \( 91.19 \pm 3.07 \) & \( \textbf{84.53} \pm \textbf{2.55} \) & \( \textbf{75.52} \pm \textbf{3.44} \) & \( \textbf{80.16} \pm \textbf{6.54} \) & \( \textbf{75.23} \pm 3.09 \) & \( \textbf{70.58} \pm \textbf{2.48} \) \\
            \bottomrule
        \end{tabularx}
        
    \end{threeparttable}
\end{table}

\textbf{Dirichlet-based Non-IID} This more challenging setup is introduced by \cite{hsu2019measuring}, which incorporates the Dirichlet distribution modulated by the concentration parameter \( \alpha \). A scenario where \( \alpha \to 0 \) means each client only sees data from a single class, while \( \alpha \to \infty \) ensures that the samples are split uniformly across clients. We apply this to the MNIST and CIFAR10 with $ \alpha  = 0.05, 0.30, 0.60$. We expanded our dataset repertoire to include larger datasets, e.g., CIFAR100 \cite{articleCIFAR} with $C = 100$ and TinyImageNet\cite{Le2015TinyIV} with $C = 200$. For these larger datasets, we set $\alpha = 1, 10, 100$ using ResNet \cite{he2015deep} for CIFAR100 and MobileNet \cite{howard2017mobilenets} for TinyImageNet as shown in \cite{Hahn2022}.

Tables \ref{tab:table2} and \ref{tab:table3}  are presented in this section. The first table provides results for the MNIST and CIFAR10 datasets, while the second focuses on CIFAR100 and TinyImageNet for top-1 acc.
\texttt{FedLEx} variants outperform the traditional variants across all values of \( \alpha \) and client numbers \( C \).  
The results for CIFAR100 and TinyImageNet show that the \texttt{FedLEx} variants again outperform across different \( \alpha \) values. This underscores the capability of \texttt{FedLEx} to deliver enhanced performance, even when applied to larger neural networks and more extensive datasets. 

%

\begin{table}[t]
    \centering
    \scriptsize
    \begin{threeparttable}
        \caption{
Results for Dirichlet non-IID setting on CIFAR100 and TinyImageNet, showing top-1 accuracy (\%) with standard deviation.}
        \label{tab:table3}
        \begin{tabularx}{\linewidth}{@{} l *{6}{C} @{}}
            \toprule
            Dataset & \multicolumn{3}{c}{CIFAR100} & \multicolumn{3}{c}{TinyImageNet} \\
            \cmidrule(lr){2-4} \cmidrule(lr){5-7}
             C & \multicolumn{3}{c}{100} & \multicolumn{3}{c}{200} \\
            \( \alpha \) & 1 & 10 & 100 & 1 & 10 & 100 \\
            \midrule

            FedAvgM\cite{hsu2019measuring} & \(54.50 \pm 1.50\) & \(61.00 \pm 3.00\) & \(76.50 \pm 6.00\) & \(40.00 \pm 3.50\) & \(52.00 \pm 1.50\) & \(63.00 \pm 7.00\) \\

            FedSgd\cite{mcmahan2023communicationefficient} & \(43.11\pm1.08\) & \(54.49\pm3.33\) & \(60.03\pm6.51\) & \(39.00\pm3.84\) & \(50.68\pm1.01\) & \(61.88\pm 8.01\) \\            
      
            FedOpt \cite{reddi2021adaptive} & \(56.62\pm1.31\) & \(63.49\pm3.94\) & \(74.82\pm8.23\) & \(43.65\pm4.23\) & \(53.51\pm3.98\) & \(66.34\pm5.32\) \\          
            
            FedProx\cite{li2020federated} & \(57.41\pm5.73\) & \(65.17\pm2.30\) & \(74.87\pm5.71\) & \(45.09\pm4.64\) & \(56.68\pm2.03\) & \(69.86\pm5.01\) \\
            \midrule

            FedLExAvgM & \(65.75 \pm 2.00\) & \(74.50 \pm 4.00\) & \(81.00 \pm 5.00\) & \(53.50 \pm 3.75\) & \(64.00 \pm 2.00\) & \(74.50 \pm 6.50\) \\

            FedLExSgd & \(65.01\pm4.63\) & \(70.10\pm5.23\) & \(80.73\pm4.71\) & \(49.45\pm4.44\) & \(58.22\pm2.39\) & \(77.09\pm3.66\) \\
            
            FedLExOpt & \(68.12\pm2.60\) & \(70.19\pm1.12\) & \(81.17\pm3.16\) & \(52.27\pm3.13\) & \(65.21\pm3.34\) & \(80.90\pm5.16\) \\

            FedLExProx & \(\textbf{70.39}\pm\textbf{1.22}\) & \(\textbf{72.54}\pm\textbf{1.92}\) & \(\textbf{85.18}\pm\textbf{3.66}\) & \(\textbf{55.50}\pm\textbf{1.82}\) & \(\textbf{66.45}\pm\textbf{3.89}\) & \(\textbf{83.10}\pm\textbf{4.44}\) \\
            \midrule

        \end{tabularx}
    \end{threeparttable}
\end{table}

\begin{table*}[t]
\centering
\scriptsize
\begin{threeparttable}
\caption{Dirichlet non-IID results for MNIST and CIFAR10, showing top-1 accuracy (\%) and standard deviation..}
\label{tab:table2}
\begin{tabularx}{\linewidth}{@{} l *{12}{C} @{}}
\toprule
Dataset & \multicolumn{6}{c}{MNIST} & \multicolumn{6}{c}{CIFAR10} \\
\cmidrule(lr){2-7} \cmidrule(lr){8-13}
$C$ & \multicolumn{3}{c}{20} & \multicolumn{3}{c}{100} & \multicolumn{3}{c}{20} & \multicolumn{3}{c}{100} \\
\cmidrule(lr){2-4} \cmidrule(lr){5-7} \cmidrule(lr){8-10} \cmidrule(lr){11-13}
\( \alpha \) & $0.05$ & $0.30$ & $0.60$ & $0.05$ & $0.30$ & $0.60$ & $0.05$ & $0.30$ & $0.60$ & $0.05$ & $0.30$ & $0.60$ \\
\midrule

FedAvgM\cite{hsu2019measuring} & \(57.50\pm7.00\) & \(71.00\pm7.20\) & \(77.00\pm4.50\) & \(45.00\pm5.00\) & \(58.00\pm5.10\) & \(65.00\pm3.50\) & \(28.00\pm3.20\) & \(38.00\pm5.30\) & \(44.00\pm2.10\) & \(16.00\pm3.50\) & \(25.00\pm4.40\) & \(33.00\pm5.60\) \\

FedSgd\cite{mcmahan2023communicationefficient} & \(48.12\pm6.67\) & \(55.32\pm6.98\) & \(60.8\pm4.25\) & \(33.36\pm4.85\) & \(36.23\pm4.90\) & \(47.50\pm3.38\) & \(18.67\pm3.05\) & \(27.79\pm5.21\) & \(32.55\pm2.07\) & \(11.45\pm3.48\) & \(15.04\pm4.32\) & \(22.13\pm5.54\) \\

FedOpt \cite{reddi2021adaptive} & \(69.53\pm5.20\) & \(80.07\pm1.56\) & \(84.08\pm2.73\) & \(50.39\pm2.56\) & \(63.80\pm1.88\) & \(70.35\pm2.41\) & \(22.55\pm3.36\) & \(41.90\pm4.33\) & \(51.36\pm2.79\) & \(23.03\pm2.58\) & \(36.87\pm2.49\) & \(40.51\pm2.69\) \\

FedProx \cite{li2020federated} & \(72.95\pm9.43\) & \(85.76\pm3.59\) & \(89.23\pm4.25\) & \(58.80\pm1.29\) & \(66.46\pm1.25\) & \(78.48\pm2.25\) & \(24.57\pm6.88\) & \(47.48\pm4.02\) & \(52.02\pm1.02\) & \(20.07\pm2.93\) & \(35.22\pm4.89\) & \(42.45\pm1.94\) \\

\midrule

FedLExAvgM & \(64.00\pm6.30\) & \(82.00\pm7.50\) & \(83.50\pm4.70\) & \(58.50\pm5.20\) & \(69.50\pm5.30\) & \(76.00\pm3.60\) & \(29.00\pm3.30\) & \(39.00\pm5.40\) & \(45.00\pm2.20\) & \(17.00\pm3.60\) & \(36.00\pm4.50\) & \(44.60\pm5.70\) \\

FedLExSgd & \(55.39\pm4.99\) & \(56.71\pm4.90\) & \(75.36\pm6.70\) & \(40.01\pm3.61\) & \(41.35\pm4.78\) & \(51.37\pm3.52\) & \(26.19\pm4.26\) & \(33.37\pm4.01\) & \(38.05\pm3.37\) & \(14.45\pm5.26\) & \(21.13\pm2.28\) & \(27.01\pm3.06\) \\

FedLExOpt & \(73.64\pm4.51\) & \(\textbf{89.84}\pm\textbf{5.39}\) & \(\textbf{91.42}\pm\textbf{3.56}\) & \(65.65\pm\ 1.41\) & \(\textbf{74.49}\pm\textbf{1.47}\) & \(\textbf{85.98}\pm\textbf{1.36}\) & \(37.35\pm3.80\) & \(55.52\pm2.20\) & \(\textbf{60.46}\pm\textbf{2.43}\) & \(30.64\pm1.18\) & \(\textbf{51.98}\pm\textbf{1.27}\) & \(54.09\pm1.12\) \\

FedLExProx & \(\textbf{79.85}\pm6.85\) & \(88.81\pm1.68\) & \(96.09\pm2.94\) & \(\textbf{77.92}\pm\textbf{1.43}\) & \(73.94\pm1.75\) & \(85.41\pm1.29\) & \(\textbf{40.07}\pm \textbf{4.42}\) & \(\textbf{56.95}\pm\textbf{2.38}\) & \(59.29\pm2.49\) & \(\textbf{33.67}\pm\textbf{2.48}\) & \(51.75\pm1.15\) & \(\textbf{55.56}\pm\textbf{1.92}\) \\
\bottomrule
\end{tabularx}
\end{threeparttable}
\end{table*}

\textbf{LEAF Non-IID} Using the LEAF benchmark \cite{caldas2019leaf}, we choose the FEMNIST and Shakespeare datasets to simulate real-world FL scenarios, with each dataset tailored for specific tasks. The FEMNIST dataset is for multiclassification involving 62 classes, and the Shakespeare data set is for the prediction of the next character with 80 characters given a sentence. We used the FEMNIST-CNN architecture shown in \cite{caldas2019leaf} on the FEMNIST dataset, and a two-layer LSTM, as proposed in \cite{mcmahan2023communicationefficient}, on the Shakespeare dataset. The structure of the Shakespeare dataset suggests to employ \( C = 730 \) and \( C = 660 \) for FEMNIST as shown in \cite{Hahn2022}.

The experimental results for top-1 accuracy for the FEMNIST and Shakespeare datasets are presented in Table \ref{tab:table3}.
For both datasets, \texttt{FedLEx} variants outperform the traditional version, indicating an improvement for a more expansive and realistic FL scenario.


\begin{table}[t]
    \centering
    \scriptsize
    \begin{threeparttable}
        \caption{
Results for LEAF non-IID setting on FEMNIST and Shakespeare, detailing top-1 accuracy  (\%) with standard deviation.}
        \label{tab:table3}
        \begin{tabularx}{\linewidth}{@{} l *{2}{C} @{}}
            \toprule
            Dataset & \multicolumn{1}{c}{FEMNIST} & \multicolumn{1}{c}{Shakespeare} \\
            \cmidrule(lr){2-2} \cmidrule(lr){3-3}
             & Acc-1 & Acc-1 \\
            \midrule
            FedAvgM\cite{hsu2019measuring} & \(69.50\pm4.00\) & \(52.50\pm6.00\) \\

            FedSgd\cite{mcmahan2023communicationefficient} & \(37.91\pm3.02\) & \(31.77\pm5.68\) \\
            
            FedAOpt \cite{reddi2021adaptive} & \(71.09\pm5.87\) & \(53.77\pm6.92\) \\
            
            FedProx \cite{li2020federated} & \(73.37\pm5.21\) & \(58.10\pm3.88\) \\

            \midrule
            FedLExAvgM & \(80.00\pm5.50\) & \(54.00\pm5.00\) \\

            FedLExSgd & \(40.09\pm7.77\) & \(35.35\pm5.51\) \\
            
            FedLExOpt & \(85.91\pm4.67\) & \(56.02\pm3.11\) \\
            
            FedLExProx & \(\textbf{89.25}\pm\textbf{1.56}\) & \(\textbf{62.10}\pm\textbf{2.05}\) \\
            \bottomrule
        \end{tabularx}
    \end{threeparttable}
\end{table}

\begin{figure}[t]
    \centering
    \includegraphics[width=\linewidth]{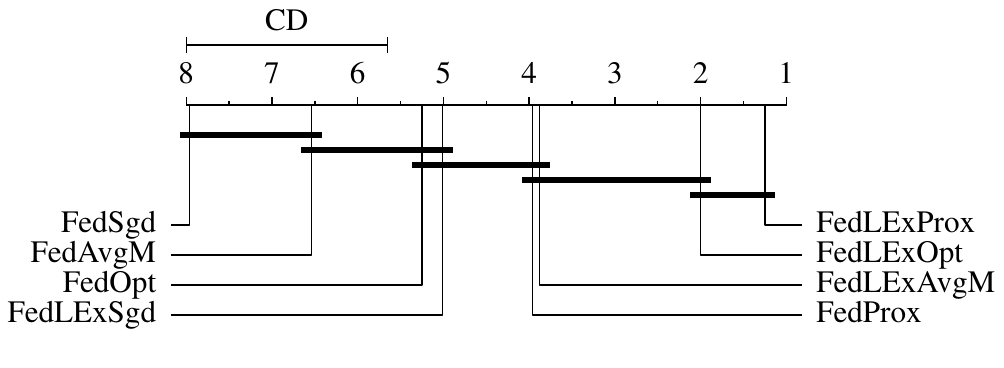}
    \caption{Critical Distance diagram summarizing the algorithms performance.}
    \label{fig:CD}
\end{figure}

\textbf{Statistical Test} We evaluate the statistical significance of our results using the Critical Distance (CD) plot, which provides a visual interpretation of the Friedman non-parametric test at a confidence level of 0.05, as described by \cite{JMLRv7demsar06a}. This was supplemented with the Nemenyi post-hoc analysis. Figure \ref{fig:CD} presents the CD plot derived from our experimental data, indicating that the \texttt{FedLex} variants consistently outperformed their counterparts in all scenarios tested, with \texttt{ FedLExProx} emerging as the best performer. 

\subsection{Ablation Studies}
Through ablation studies, we dissect the core components of the \texttt{FedLEx} to determine the significance of each hyperparameter and its impact on overall performance. These insights are instrumental in tuning \texttt{FedLEx} for diverse FL scenarios. 
The number of explorer clients \( C_{exp} \) influences the comprehensiveness of \( G_{\text{global}} \) in capturing the intricacies of the clients loss landscapes. We conducted experiments with $C_{exp} = {1, 0.75, 0.50, 0.25}$ as percentage. We find that even a relative modest \( C_{exp} \) value can lead to significant performance enhancements.
The depth of the exploration phase, quantified by the number of exploration epochs \( E_{exp} \) influences loss function surface understanding and \( G_{\text{global}} \)’s knowledge depth. We evaluated FedLEx with \( E_{exp} = {150, 300, 500, 750, 1000}\).

\begin{figure*}[t]
    \centering
    \includegraphics[width=\linewidth]{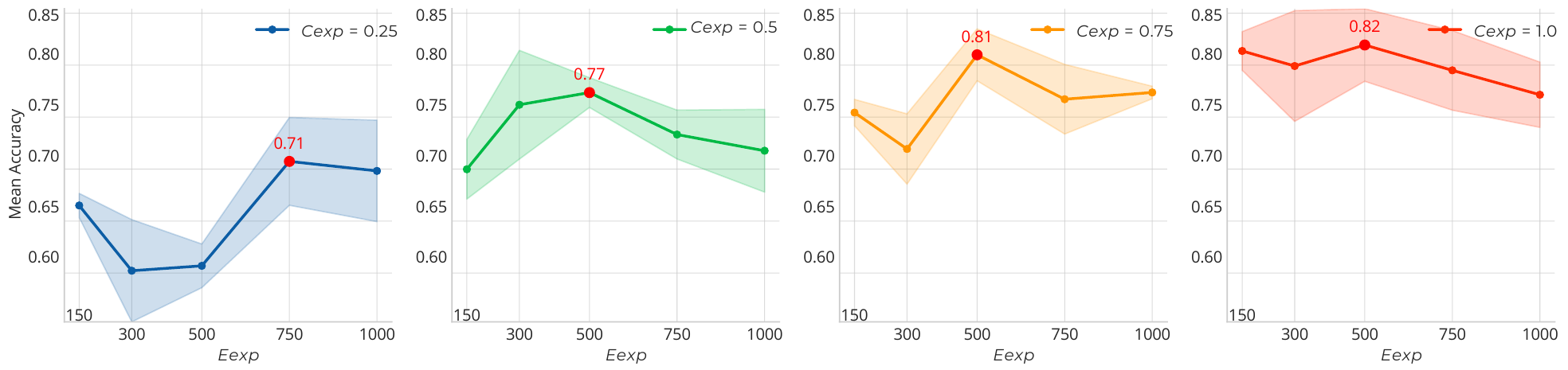}
    \caption{Every graph shows the accuracy and standard deviation over different 
    $E_{exp}$ and $C_{exp}$ ratios, highlighting optimal performance.}
    \label{fig:ablation}
\end{figure*}

It is possible to observe in Figure \ref{fig:ablation} the trend of top-1 accuracy and deviation of the combination of ablation experiments in the MNIST data set with a pathological non-IID setting, with $R = 100$ and $C = 20$. The x-axis represents the \( E_{exp} \) parameter. Each distinct plot corresponds to a different \( C_{exp} \), with the highest accuracy achieved distinctly highlighted.
In particular, our findings reveal that even a conservative value of $C_{exp}$ can boost the accuracy. The influence of $C_{exp}$ is substantial, with higher counts of explorer clients resulting in improved initial accuracy, indicative of a more robust \( G_{\text{global}}\) at the outset of the learning process. A discernible trend suggests that increasing the $E_{exp}$ value generally leads to an improvement in accuracy. However, at higher values, the increase is reduced, indicating a saturation point beyond which additional exploration epochs no longer improve accuracy.

\section{Conclusion and Future Work}
In this paper, we presented \texttt{FedLEx}, a novel and scalable aggregation method designed to improve FL in non-IID scenarios while maintaining privacy requirements. \texttt{FedLEx} enhances the global model performance by aggregating and sharing the knowledge of the client's gradient in various loss landscapes.
This mapping guides and scales the gradient in subsequent training sessions, targeting superior updates from clients for the global model.

Central to \texttt{FedLEx} is the creation and sharing of the global guidance matrix containing the gradient guidance parameters. These parameters result from a collaborative exploration of the loss surfaces. This approach, combined with the collaborative essence of FL, optimizes the gradient, thus enabling faster convergence of the global model in challenging non-IID environments. Our experiments have shown that \texttt{FedLEx} is a versatile and effective schema. Incorporating \texttt{FedLEx} into FL algorithms significantly improves performance.

Future work for \texttt{FedLEx} should focus on developing algorithms for optimizing the global guidance matrix based on data set specifics, model conditions, or creating multiple matrices for varied client groups. It is vital to investigate its scalability in large neural networks with numerous clients, considering the latency and bandwidth. Additionally, testing \texttt{FedLEx} in Federated Class-Continual Learning scenarios with dynamically added new classes and enhancing its scalability and robustness against adversarial attacks and data poisoning are key areas for further research.

\bibliographystyle{IEEEtran}
\bibliography{bibliography.bib}

\end{document}